\def\year{2021}\relax
\documentclass[letterpaper]{article} 
\usepackage{aaai21}  
\usepackage{times}  
\usepackage{helvet} 
\usepackage{courier}  
\usepackage[hyphens]{url}  
\usepackage{graphicx} 
\usepackage{natbib}  
\usepackage{caption} 

\usepackage{latexsym}
\usepackage{multirow}
\usepackage{booktabs}
\usepackage{nicefrac}
\usepackage{subcaption}
\usepackage{bbm}
\usepackage{amsmath,amsfonts,amsthm,amssymb}
\usepackage{microtype}

\newcommand{\tabincell}[2]{\begin{tabular}{@{}#1@{}}#2\end{tabular}}
\graphicspath{{./imgs/}}

\title{Converse, Focus and Guess - Towards Multi-Document Driven Dialogue}
\author{Han Liu\textsuperscript{\rm 1}, Caixia Yuan\textsuperscript{\rm 1}\thanks{Corresponding author.}, Xiaojie Wang\textsuperscript{\rm 1},\\
  Yushu Yang\textsuperscript{\rm 2}, Huixing Jiang\textsuperscript{\rm 2}, Zhongyuan Wang\textsuperscript{\rm 2} \\
}

\affiliations{
    \textsuperscript{\rm 1}Beijing University of Posts and Telecommunications, \textsuperscript{\rm 2}Meituan Group \\
    \{liuhan,yuancx,xjwang\}@bupt.edu.cn, \{yangyushu, jianghuixing, wangzhongyuan02\}@meituan.com
}

\begin{document}
\maketitle

\begin{abstract}
    We propose a novel task, \textbf{M}ulti-\textbf{D}ocument \textbf{D}riven \textbf{D}ialogue (\textbf{MD3}), in which an agent can guess the target document that the user is interested in by leading a dialogue. To benchmark progress, we introduce a new dataset of GuessMovie, which contains 16,881 documents, each describing a movie, and associated 13,434 dialogues. Further, we propose the MD3 model. Keeping guessing the target document in mind, it converses with the user conditioned on both document engagement and user feedback. In order to incorporate large-scale external documents into the dialogue, it pretrains a document representation which is sensitive to attributes it talks about an object. Then it tracks dialogue state by detecting evolvement of document belief and attribute belief, and finally optimizes dialogue policy in principle of entropy decreasing and reward increasing, which is expected to successfully guess the user's target in a minimum number of turns. Experiments show that our method significantly outperforms several strong baseline methods and is very close to human's performance.
    \footnote{\url{https://github.com/laddie132/MD3}}
\end{abstract}

\section{Introduction}
The recent progress with human-machine dialogue techniques enable conversational agents to be extensively applied in customer service, information retrieval, personal assistance and so on. In order to assist the user to accomplish specific tasks, the agent must necessarily query external knowledge. Several works have focused on incorporating structured knowledge base (KB) into dialogues \citep{dhingra2017towards,madotto2018mem2seq,wu2019global} through KB lookup. Although these efforts scales nicely to huge knowledge base, many real-world task-oriented dialogues involve in referring to a great number of documents (such as manuals, instruction booklets, and other informational documents). Since the complexity of document understanding, developing dialogues with many grounding articles is far from a trial task. 

\begin{figure*}[htb]
    \centering
    \includegraphics[width=0.9\textwidth]{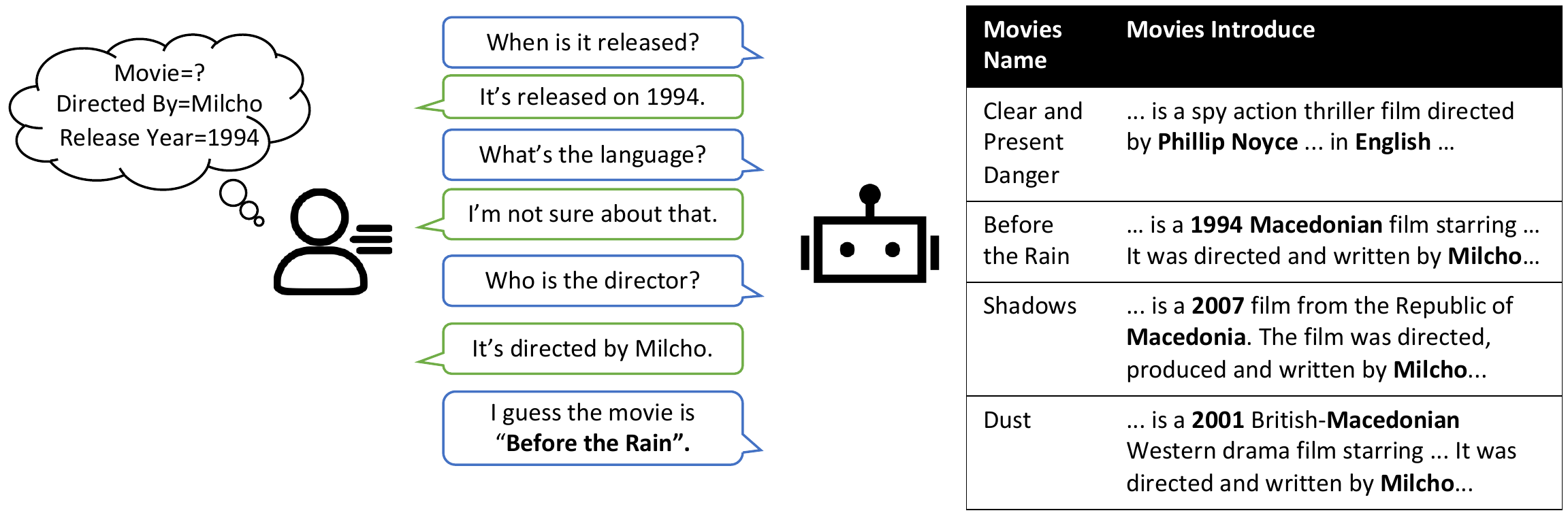}
    \caption{Sample dialogue from GuessMovie dataset. The user's target is ``\textit{Before the Rain}'' and only knows its director is Milcho, and release year is \textit{1994}. The goal of the agent is to correctly guess the target movie via asking minimum number of questions to the user. Here we only demonstrate four candidate movie documents due to space constraints (any number of documents could be provided in practice). Since candidate movies have the highest uncertainty (i.e. the largest information entropy) on ``\textit{release year}'', the agent ask ``\textit{When is it released?}'' to minimize the range of candidates. After the first turn, the last two movies ``\textit{Shadows}'' and ``\textit{Dust}'' can be excluded. Similarly, the agent ask ``\textit{in language}'' and ``\textit{directed by}'' in the second and third turn.
    }
    \label{fig:exp}
\end{figure*}

In this paper, we consider in particular the problem of multi-document driven dialogue (MD3), where the agent leads a dialogue with the user with a particular conversation goal and the engagement of multiple documents. To this end, we propose a MD3 game---\textit{GuessMovie}. In the game, the user selects a movie at the beginning of dialogue, which is unknown to the agent. The agent is provided with a set of candidate documents, each describing a movie, and tries to guess which movie the user selects by asking a series questions (e.g. ``\textit{Who is the director of the movie?}'' or ``\textit{When is it released?}''). The user informs the agent his answer or says ``\textit{unknown}''. The agent's goal is to guess the target movie in the shortest dialogue turns. It assumes that only the agent has access to the documents, and at least one document describing the target movie. Figure \ref{fig:exp} shows an example.

Two key challenges arising from MD3 task is: (1) how to efficiently encode and incorporate a large scale of long unstructured documents in a dialogue, and (2) how to learn optimal dialogue policy to efficiently fulfill dialogue task with smooth engagement of external documents.

To address the first challenge, we propose an factor aware document embedding, which assumes that a text is typically composed of several key factors it wants to narrate. For example, a text describing a movie mainly consists of attributes like ``\textit{directed by}'', ``\textit{release year}'', ``\textit{genre}'', and so on. For each attribute, attribute aware document embedding is trained with the goal of correctly recognizing the document containing the given attribute and attribute value from a set of documents. The document is finally represented as the concatenation of all attribute aware document embeddings. 

In order to manipulate the dialogue towards efficiently guessing the correct target document, the proposed MD3 model traces dialogue state by monitoring a document belief distribution and attribute belief distribution, both in global dialogue-level and temporal turn-level, which is viewed an explicit representation of dialogue dynamics leading to target movie. Besides, the model also calculates the uncertainty of each attribute through a document differentiated representation. In order to fulfill the dialogue goal within minimum dialogue turns, at each turn the policy model tends to discussing the attribute with the highest belief and highest uncertainty with \textit{asking} the user. When the belief of a document is higher than a threshold, the agent will execute a \textit{guess} action. A dialogue is deemed as ended up with a \textit{guess} action.

Remarkably, the above dialogue state tracing and dialogue policy optimizing are jointly trained using reinforcement learning. 

Our contributions are summarized as follows:
\begin{enumerate}
    \item We propose a novel MD3 task, and release a new publicly-available benchmark dialogue corpus---GuessMovie, that we hope will help further work on document-driven task-oriented dialogue agents.
    \item We introduce the MD3 model, a highly performant neural dialogue agent that is able to smoothly incorporate multiple documents through entangling document representation, document belief and attribute belief to the dynamics of the dialogue. As far as we know, this is the first study of task-oriented dialogue based on a large scale of documents.
    \item The proposed document-aware dialogue policy achieve the maximum dialogue success rate in the minimum number of dialogue turns compared with several baseline models.
\end{enumerate}

\section{Related Work}
The closest work to ours lies in the area of dialogue system incorporating unstructured knowledge. \citet{ghazvininejad2018knowledge,parthasarathi2018extending} use an Encoder-Decoder architecture where the decoder receives as input an encoding of both the context utterance and the external text knowledge. \citet{dinan2018wizard,li2019incremental} investigate extended Transformer architecture to handle the document knowledge in multi-turn dialogue. \citet{reddy2019coqa} use documents as knowledge sources for conversational Question-Answering. This line of work aims either at producing more contentful and diverse responses, or at extracting answers of user questions. The dialogue agent we build has a specific goal throughout the dialogue, from this point of view the dialogue system in this paper is task-oriented one. While these works mainly focusing on chatting about the content of a given document without a specific dialogue goal. Besides, our task differs from them in that our agent interact with a large-scale external documents, which poses new challenges for grounding dialogues.

Another line of related work is on ``\textit{Guess}''-style dialogues, among which \textit{Q20 Game} \citep{burgener2006artificial,zhao2016towards,hu2018playing} is a typical object guessing game. In Q20, the agent guesses the target object within 20 turns of questions and answers. Each object is tied with a structured KB and the user is restricted to passively answer ``\textit{Yes, No or Unknown}''. \citet{dhingra2017towards} proposes a new task and method named KB-InfoBot, which can be regarded as an extension of Q20 Game. It aims to retrieval from the structured KB through a dialogue by a soft query operation. These works mainly focus on integrating structured knowledge into dialogue systems, while it requires a lot of work to build up, and is only limited to expressing precise facts. Documents as knowledge are much easier to obtain and provide a wide spectrum of knowledge, including factoid, event logics, subjective opinion, etc. 

In the field of computer vision, both the ImageGuessing \citep{das2017learning} and GuessWhat \citep{de2017guesswhat} try to guess a picture or an object through multi-turn dialogue, which greatly expands the range of dialogue applications \citep{pang_2020_AAAI,pang_2020_ECCV}. We use a similar approach to extend dialogue games into the document-driven ones. Different from the vision field, how to encode large-scale text information is a vital challenge.

\section{Dataset: GuessMovie}
We build a benchmark \textit{GuessMovie} dataset for MD3 task on the base of the dataset WikiMovies \citep{miller2016key}. In WikiMovies, there is a large-scale document knowledge. Each is a movie introduction text, which is derived from the first paragraph of Wikipedia. In addition, it also contains a structured KB of the same movie collection, which is derived from MovieLens, OMDB and other datasets. There are totally 10 different attributes in movie KBs. But we select the common 6 ones since others are randomly discussed in the corresponding documents.

As for GuessMovie, we firstly align a document with a piece of structured knowledge. For a specific movie, some attribute might be missing from the text and the KB, and some attribute might hold more than one values (e.g. a movie may have more than one starred actors.). Such cases are preserved in GuessMovie, which is expected to make the dialogue more diverse and realistic.

\begin{table*}[htb]
    \centering
    \begin{tabular}{l|cccccc}
        \hline
        Attr & \tabincell{c}{directed by} & \tabincell{c}{release year} & \tabincell{c}{written by} & \tabincell{c}{starred actors} & \tabincell{c}{has genre} & \tabincell{c}{in language} \\ \hline
        {Num} & 14853 & 16299 & 12712 & 13204 & 12118 & 3071 \\
        {Ent} & 6187 & 103 & 10404 & 10180 & 23 & 96 \\
        {Ave} & 1.05 & 1.00 & 1.48 & 2.48 & 1.27 & 1.10 \\
        \hline
    \end{tabular}
    \caption{GuessMovie dataset statistics. \textit{Num} denotes the number of documents containing it (with total 16,881 documents). \textit{Ent} denotes the number of rare values. \textit{Ave} denotes the average number of values that an attribute has.}
    \label{tab:data-attr}
\end{table*}

Then we create multi-turn dialogues for some documents. We design a dialogue simulator that interacts with the structured KB to generate dialogues. It consists of two agents playing the roles of the user and the system. Both agents interact with each other using a finite set of dialogue acts directing the dialogue. The user simulator is constructed in a handcrafted way introduced in the following section. The system agent is provided with a candidate KB, including the target knowledge and a set of other randomly selected knowledge. At the beginning, the system agent generates a dialogue act of ``\textit{asking}'' an attribute (e.g. ``\textit{when is the movie released?}''). The probability of an attribute being chosen as ``\textit{asked attribute}'' is proportional to the information entropy computed according to its distribution within the candidate KB. After a turn of ``question-answering'', the KBs insistent with the facts so far are filtered out. The dialogue continues until the system agent is confident with the target KB and executes a ``\textit{guess}'' action. For natural language generation, we use several diverse natural language patterns that takes an attribute or an attribute-value pair as argument. 

Totally, GuessMovie is comprised up with 13,434 dialogues for 16,881 documents. The average length of documents is 107.66 words. More statistics are described in Table \ref{tab:data-attr}.

It's worth noting that we do not include any domain-specific constraints in both simulated agents. Although our examples use Wikipedia articles about movies, we see the same techniques being valid for other external documents such as manuals, instruction booklets, and other informational documents, as far as they can be loosely structed as several facets about an object.

\section{Method}
\begin{figure*}[htb]
    \centering
    \includegraphics[width=0.9\textwidth]{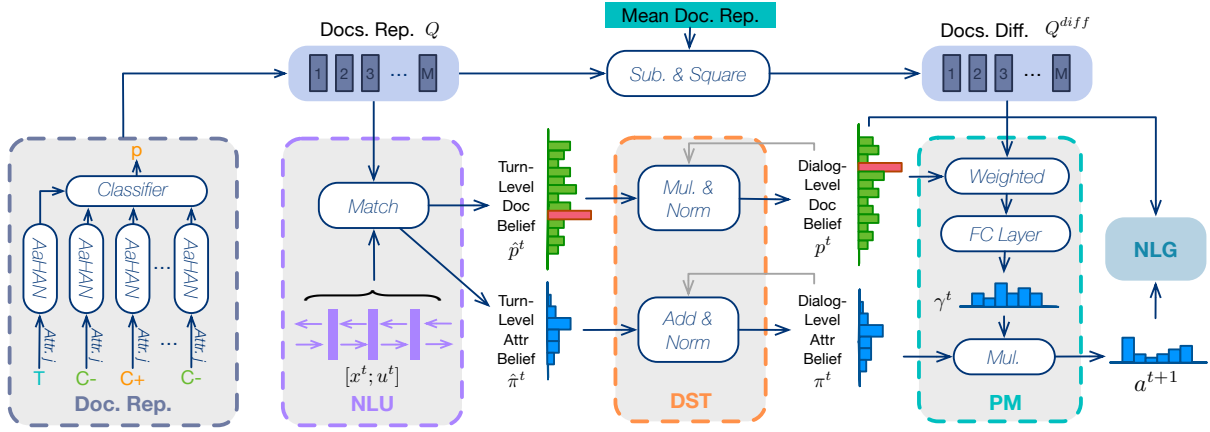}
    \caption{Overall architecture of MD3 model with AaDR for document representation, DaLU for NLU, DaST for DST, DaPO for PM and rule for NLG.}
    \label{fig:model}
\end{figure*}

Figure \ref{fig:model} illustrates the overall architecture of the \textit{\textbf{M}ulti-\textbf{D}ocument \textbf{D}riven \textbf{D}ialogue (\textbf{MD3})} model, including five parts: document representation, Natural Language Understanding (NLU), Dialogue State Tracking (DST), Policy Model (PM) and Natural Language Generation (NLG). This section introduces each part in details.

\paragraph{Task Definition} Given a set of $M$ documents $\{D_i\}_{i=1}^M$ as candidates, and a target document known only by the user (each of which corresponds to a unique object, i.e. a movie in our case), the agent outputs a consecutive of responses that ask an attribute (e.g., ``\textit{when is it released?}'', ``\textit{Is it released in 1990?}''), or guesses a target document and ends the dialogue. The user can provide the answer to the questions, or answer like ``\textit{I don't know}''.

\subsection{AaDR for Document Representation} \label{sec:doc-pt}
As for encoding large-scale documents along with optimizing dialogue agent is computationally burdening and inviable, we resort to pretrain the documents in advance. We argue that the pre-trained representation should be not only liable to the original meaning but also helpful for constructing a document-driven task-oriented dialogue agent. 

To this end, we propose \textit{\textbf{A}ttribute-\textbf{a}ware \textbf{D}ocument \textbf{R}epresentation (AaDR)}. We assume that each document talks about several attributes with index $\{1,...,j,...,L\}$, such as ``\textit{directed by}'', ``\textit{release year}'', ``\textit{in language}'' and so on in movie scenario.

Inspired by Hierarchical Attention Networks (HAN) \citep{yang2016hierarchical} and Hierarchical Label-Wise LSTM \citep{liu2020label}, we introduce an \textit{Attribute-aware HAN (AaHAN)} to encode each document, which seen attribute as label. Specifically, each attribute is used to index the corresponding parameters in hierarchical attention. This mechanism can capture different information for different attributes. Combined with the contrastive loss \citep{hadsell2006dimensionality}, an attribute-aware document representation is learned.

\paragraph{Pre-training}
Randomly sample a target document as $ T $. For an attribute-value pair $(a_{j}, v_{jk})$, we sample a positive document $C^{+}$ which has the same value $v_{jk}$ for attribute $a_j$ as $T$ and several negatives $C^{-}$ which have different attributes values. These two parts are combined to a training sample $\{C_i\}_{i=1}^R$ for $a_{j}$. The training goal is to distinguish the positive document from all $\{C_i\}_{i=1}^R$.

As for the target $T$ and a candidate $C_i$, we can calculate the corresponding document representation on a certain attribute $a_j$: $H^{t}_j \in \mathbb{R}^{2d}$ and $H^{c_i}_{j} \in \mathbb{R}^{2d}$. Further, calculate the similarity of two documents directly as follows.
\begin{align}
    H^{t}_j &= \operatorname{AaHAN}^{t}(T, a_j) \\
    H^{c_i}_{j} &= \operatorname{AaHAN}^{c}(C_i, a_j)
\end{align}

It's trained using a negative log-likelihood loss function.
\begin{equation}
    \mathop{\mathbb{E}}\limits_{t,j,c^+,c^-}\left[-\log \frac{\exp{((H^{t}_j)^T H^{c^+}_j})}{\exp{((H^{t}_j)^T H^{c^+}_j)} + \sum_{c^-}\exp{((H^{t}_j)^T H^{c^-}_j})}\right]
\end{equation}

\paragraph{Representation}
After pre-training, for each document $D_i$, we can obtain the representation $Q_{ij} \in \mathbb{R}^{4d}$ on attribute $ a_j $, which is the concatenation of outputs of target encoder and candidate encoder.
\begin{align}
    H^{t}_{ij} &= \operatorname{AaHAN}^{t}(D_i, a_j) \\
    H^{c}_{ij} &= \operatorname{AaHAN}^{c}(D_i, a_j) \\
    Q_{ij} &= [H^{t}_{ij};H^{c}_{ij}]
\end{align}

After that, the final document representation $Q_i \in \mathbb{R}^{L \times 4d}$ is obtained by concatenating $L$ attribute-aware representation.

\subsection{DaLU for NLU}
We propose \textit{\textbf{D}ocument-\textbf{a}ware \textbf{L}anguage \textbf{U}nderstanding (DaLU)} for NLU. In the previous turn, the agent's question is $ x ^ {t} $, and the user's response is $ u ^ {t} $, which are together concatenated into a long sentence and encoded using a BiGRU. Take the last hidden state as output $G^t \in \mathbb{R}^{2d}$.

Therefore, the similarity $\hat{S}^t \in \mathbb{R}^{M \times L}$ between previous turn $G^t \in \mathbb{R}^{2d}$ and candidate documents $Q \in \mathbb{R}^{M \times L \times 4d}$ (concatenation of each candidate $Q_i$ etc.) can be calculated directly by a bilinear method. It reflects the matching degree for each candidate $Q$.
\begin{equation}
    \hat{S}^t = G^t W^s Q^T
\end{equation}
where $W^s \in \mathbb{R}^{2d \times 4d}$ is a trainable parameter.

In addition, we further calculate two distributions: (1) the attribute type $\tilde{\pi}^t \in \mathbb{R}^{L}$. (2) a flag $\alpha^t \in \mathbb{R}^1$ indicating whether the response is ``\textit{unknown}''. The closer the value is to 1, the less valid information is included in this turn.
\begin{align}
    \tilde{\pi}^t &= \operatorname{softmax}(W^{attr}G^t) \\
    \alpha^t &= \operatorname{sigmoid}(W^{unk}G^t)
\end{align}
where $W^{attr} \in \mathbb{R}^{L \times 2d}$ and $W^{unk} \in \mathbb{R}^{1 \times 2d}$ are trainable parameters.

When previous response is ``\textit{unknown}'', the selected probability of each candidate is equal on turn level. Therefore, we expand the similarity $\hat{S}^t$ on attribute dimension, concatenating a vector whose similarity is all $1$, and get $S^t \in \mathbb{R}^{M \times (L+1)}$. For the attribute type $\tilde{\pi}^t$, we also expand the attribute dimension by fusing the $\tilde{\pi}^t$ and $\alpha^t$, and get $\beta^t \in \mathbb{R}^{(L+1)}$. 
\begin{align}
    S^t &= [\hat{S}^t;\mathbbm{1}] \\
    \beta^t &= [\tilde{\pi}^t(1 - \alpha^t);\alpha^t]
\end{align}

Further the Turn-Level Doc \footnote{Doc is the abbreviation of document and Attr is attribute.} Belief $\hat{p}^t \in \mathbb{R}^{M}$ is obtained, which is the probability of each candidate document being selected at current turn, and also the Turn-Level Attr Belief $\hat{\pi}^t \in \mathbb{R}^L$.
\begin{align}
    \hat{p}^t &= \operatorname{softmax}(S^t \beta^t) \\
    \hat{\pi}^t &= \alpha^t \tilde{\pi}^t
\end{align}

\subsection{DaST for DST}
We propose \textit{\textbf{D}ocument-\textbf{a}ware \textbf{S}tate \textbf{T}racking (DaST)} for DST. The \textit{dialogue state} is defined as the following two parts: (1) Dialog-Level Doc Belief $ p^t \in \mathbb{R}^{M}$ represents the probability of each document being selected. (2) Dialog-Level Attr Belief $\pi^t \in \mathbb{R}^{L}$ is the probability of an attribute being unknown. The lower the value, the higher the attribute belief.

When a document is excluded, it will rarely be selected. But the attribute belief is accumulated for each one, indicating whether the attribute will be asked. This two belief distributions are updated as follows.
\begin{align}
    p^t &= \operatorname{norm}(p^{t-1} \circ \hat{p}^t) \\
    \pi^t &= \operatorname{min}(\pi^{t-1}+\hat{\pi}^t, 1)
\end{align}
where $\operatorname{norm}$ is the L1-Normalization method. The initial value $p^0$ is a uniform distribution, and $\pi^0$ is initialized to zero vector. At the beginning of each dialogue, agent directly use $p^0$ and $\pi^0$ into the PM module. 

\subsection{DaPO for PM}
We propose \textit{\textbf{D}ocument-\textbf{a}ware \textbf{P}olicy \textbf{O}ptimizing (DaPO)} for PM to minimize the number of dialogue turns and guess the true target.

In order to describe the degree of discreteness of the data, we introduce a similar variance measure to calculate the differentiated representation $Q^{diff}_i \in \mathbb{R}^{L \times 4d}$ for each document $Q_i$. It's used to describe the degree of attribute discreteness.
\begin{align}
    \overline{Q} &= \frac{1}{N}\sum_{i=1}^N Q_i\\
    Q^{diff}_i &= (Q_i - \overline{Q})^2
\end{align}
where $N$ is the size of whole dataset.

Note that the Dialog-Level Doc Belief $p^t$ is the confidence of each document. Therefore, a weighted sum is used on the differentiated representation $Q^{diff} \in \mathbb{R}^{M \times (L \times 4d)}$ (concatenation of each candidate $Q_i^{diff}$ etc.) to obtain $v^t \in \mathbb{R}^{L \times 4d}$, which is used to describe attribute uncertainty $\gamma^t \in \mathbb{R}^{L}$ over all document candidates.
\begin{align}
    v^t &= (Q^{diff})^T p^t \\
    \gamma^t &= v^t W^{diff}
\end{align}
where $W^{diff} \in \mathbb{R}^{4d \times 1}$ is a trainable parameter.

Normally, the agent can directly ask the attribute with the largest uncertainty $\gamma^t$, expected to successfully end the dialogue in a minimum number of turns. However, some highly uncertain attributes may have low belief, therefore, the ``\textit{ask}'' action of time step $t+1$ should be calculated as:
\begin{equation}
    a^{t+1} = \operatorname{softmax}(\gamma^t (1 - \pi^t))
\end{equation}

It is also interesting to note that $a^{t+1}$ is equivalent to the probability of each attribute being asked at timestep $t+1$.

\subsection{Rule for NLG}
In NLG module, the agent generates natural language response respectively for ``\textit{ask}'' and ``\textit{guess}'' action. For ``\textit{ask}'' action, it produces a response of asking the most probable attribute greedily according to $a^{t+1}$. For ``\textit{guess}'' action, the agent guesses the movie corresponding with the most probable document greedily according to $ p^t $. We use predefined natural language templates to converse with the user. The termination includes two cases.

\begin{enumerate}
    \setlength{\itemsep}{0pt}
    \setlength{\parsep}{0pt}
    \setlength{\parskip}{0pt}
    \item Positive termination: when the maximum value of Dialog-Level Doc Belief $p^t$ exceeds a certain threshold $ K $.
    \item Passive termination: when the set maximum number of dialogue turns is reached.
\end{enumerate}

\section{Experiments}
\subsection{Experimental Setting}
We divide GuessMovie into two disjoint parts. The 70\% part is used for pre-training document representation and NLU module, and the remaining 30\% is used for training MD3 with 50k simulations using REINFORCE \citep{williams1992simple} algorithm. The discount rate is 0.9. After training, we run a further 5k simulations to test the performance. 

We construct a user simulator \citep{schatzmann2007agenda,li2016user} with handcrafted rules because the user only need to answer the agent's questions passively. It randomly selects the target at the beginning. During the dialogue, the relevant value is indexed from the current structured KB and filled into a natural language template to response. Otherwise, if the user doesn't have knowledge, the answer is unknown. Specifically, in order to increase the dialogue diversity, we randomly mask some values for the 6 attributes with proportion of 0.1 at the beginning of each dialogue. However, all documents remain unchanged in the agent.

The reward function is similar to \citet{dhingra2017towards}. If the rank of target document is in the top $R=3$ results, the reward is $ max(0, 2 (1 - (r - 1) / R)) $, where $r$ is the actual rank of the target. Otherwise, if the dialogue failes, the reward is $-1$. In addition, a reward of $ -0.1 $ will be given in each turns, making the dialogue tend to be finshed in a smaller number of turns.

We use Adam\citep{kingma2014adam} optimizer with learning rate 0.001 and GloVe \citep{pennington2014glove} word embedding. The number of candidate documents for document representation and dialogue is 32 by default. The maximum number of turns is 5. The probability threshold $K$ for whether performing a Guess action is 0.5.

\begin{table*}[htb]
    \centering
    \small
    \begin{tabular}{ll|lllll|lllll|lllll}
        \hline
        \multirow{2}*{NLU} & \multirow{2}*{PM} & \multicolumn{5}{c|}{32} & \multicolumn{5}{c|}{64} & \multicolumn{5}{c}{128} \\ \cline{3-17}
        & & \textbf{S1} & \textbf{S3} & \textbf{M} & \textbf{T} & \textbf{R} & \textbf{S1} & \textbf{S3} & \textbf{M} & \textbf{T} & \textbf{R} & \textbf{S1} & \textbf{S3} & \textbf{M} & \textbf{T} & \textbf{R} \\ \hline
        
        MRC & Rand & .51 & .72 & .63 & 5.00 & 0.87 & .40 & .62 & .56 & 5.00 & 0.57 & .28 & .53 & .42 & 5.00 & 0.27\\
        MRC & Fixed & .67 & .89 & .78 & 5.00 & 1.29 & .57 & .79 & .69 & 5.00 & 1.06 & .49 & .67 & .61 & 5.00 & 0.76 \\ 
        MRC & DaPO & .79 & .94 & .87 & 5.00 & 1.49 & .71 & .93 & .82 & 5.00 & 1.41 & .52 & .86 & .69 & 5.00 & 1.23 \\ \hline \hline
        DaLU & Rand & .56 & .81 & .70 & 4.34 & 1.21 & .47 & .71 & .62 & 4.64 & 0.91 & .38 & .61 & .53 & 4.84 & 0.57 \\
        DaLU & Fixed & .71 & .93 & .83 & 3.78 & 1.57 & .65 & .87 & .77 & 4.28 & 1.37 & .56 & .80 & .70 & 4.66 & 1.14\\
        \textbf{DaLU} & \textbf{DaPO} & \textbf{.83} & \textbf{.97} & \textbf{.90} & \textbf{3.42} & \textbf{1.73} & \textbf{.74} & \textbf{.93} & \textbf{.84} & \textbf{3.86} & \textbf{1.56} & \textbf{.67} & \textbf{.88} & \textbf{.78} & \textbf{4.29} & \textbf{1.38}\\
        DaLU & w/o AU & .65 & .91 & .79 & 3.88 & 1.51 & .62 & .86 & .75 & 4.42 & 1.33 & .55 & .81 & .70 & 4.76 & 1.14 \\
        DaLU & w/o AB & .47 & .82 & .65 & 4.64 & 1.14 & .45 & .74 & .61 & 4.48 & 0.98 & .36 & .63 & .52 & 4.70 & 0.66 \\ \hline \hline
        Human & Human & \textbf{.99} & \textbf{.99} & \textbf{.99} & \textbf{3.28} & \textbf{1.87} & \textbf{.98} & \textbf{.99} & \textbf{.98} & \textbf{3.52} & \textbf{1.83} & \textbf{.96} & \textbf{.99} & \textbf{.97} & \textbf{3.80} & \textbf{1.79} \\ \hline        
    \end{tabular}
    \caption{Dialogue test results with various combinations of NLU and PM on GuessMovie dataset. \textit{S1} denotes the top-1 dialogue success rate. \textit{S3} denotes the top-3 dialogue success rate. \textit{M} denotes the target document ranking metric MRR. \textit{T} denotes the average number of dialogue turns. \textit{R} denotes the average rewards. \textit{AU} denotes attribute uncertainty. \textit{AB} denotes attribute belief.}
    \label{tab:results}
\end{table*}

\subsection{Baselines}
The following introduces several different modules to compare with our method, including the machine reading comprehension (MRC) for NLU, random and fixed policy for PM.

\paragraph{NLU}
Here we introduce two methods.
\begin{itemize}
    \setlength{\itemsep}{0pt}
    \item \texttt{MRC}: directly use the attributes and values for a document, which is extracted by BERT\citep{devlin2018bert} with no answer supported. The part of GuessMovie dataset used for documents pre-training are modified into a standard extractive MRC dataset, and divided into training, development and test parts. The testing EM score is 82.99 and F1 score is 87.44. After training, we extract the structured MRC-KB from the other part of GuessMovie dataset for dialogue. The similarity of each movie and the current user response is measured by calculating the overlap ratio, and normalized as the Turn-Level Doc Belief $\hat{p}^t$. The Turn-Level Attr Belief $\hat{\pi}^t$ can be obtained directly.
    \item \texttt{DaLU}: the method proposed in this paper.
\end{itemize}

\paragraph{PM}
Here we introduce three methods.
\begin{itemize}
    \item \texttt{Rand}: randomly selecting an attribute to ask.
    \item \texttt{Fixed}: asking attributes in a fixed order.
    \item \texttt{DaPO}: the method proposed in this paper.
\end{itemize}

\paragraph{Human Performance}
Human performance is the ceiling of this task by assuming human can understand the candidate documents accurately, and calculate the truth document distribution. An attribute with the most discriminating information can be generated. In this scenario, we directly adopt structured KB and attribute-value pairs instead of natural language, in order to simulate accurate interaction.
\begin{itemize}
    \item \texttt{Human-NLU}: we use a handcrafted module to directly match the current turn with each structured KB. If they match, the selection probability in Turn-Level Doc Belief $\hat{p}^t$ is set to 1, otherwise the probability is set to 0. It's the same for Turn-Level Attr Belief $\hat{\pi}^t$.
    \item \texttt{Human-PM}: the Dialog-Level Doc Belief $p$ is thresholded to obtain a filtered documents. Based on the structured KB, the entropy of each attribute is calculated, and normalized as distribution $\gamma^t$. After that, it is still fused with the Dialog-Level Attr Belief $\pi^t$.
\end{itemize}

\subsection{Results}
We simulated 5k dialogues randomly for testing the performance. The complete results are shown in the Table \ref{tab:results}. Various combinations of NLU and PM module are selected, and different size of candidate documents (\textit{32, 64, and 128}) are setted. We calculated the dialogue success rate of the target in the candidates top-1 or top-3. In addition, similar to the retrieval task, the Mean Reciprocal Rank (MRR) is also calculated based on the position of the target in the ranking result. At the same time, the average number of dialogue turns and average rewards are also obtained.

As shown in the Table \ref{tab:results}, the DaLU-NLU module has a higher dialogue success rate and smaller number of turns compared to the MRC-NLU module with the same PM, because it's difficult for the MRC model to accurately extract the attributes and there may be implied attributes value in the document. Therefore more interaction is needed to improve the select probability of the target.

In the combinations of DaLU-NLU, DaPO-PM is significantly superior to the random or fixed policy which is impossible to generate appropriate actions for specific candidates, while DaPO-PM can ask attribute with higher uncertainty and belief.

Although MD3 has significant advantages compared with others and is very close to human's performance, it's performance reduced quickly when increasing the size of candidate documents to 64 or 128. Also, there is still a lot of room for improvement on top-1 dialogue success rate and larger size of candidate documents.

\begin{figure*}[htb]
    \centering
    \subcaptionbox{TDR (MD3)}{
        \includegraphics[width=0.23\textwidth]{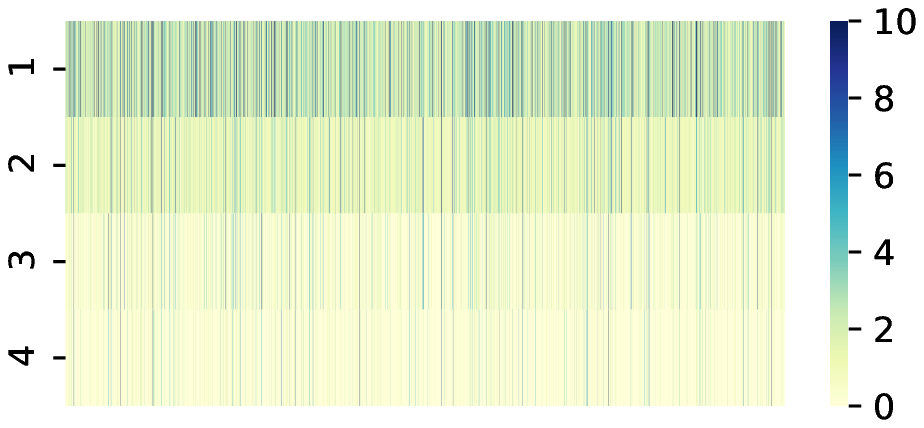}
    }
    \subcaptionbox{TDR (MD3-Rand)}{
        \includegraphics[width=0.23\textwidth]{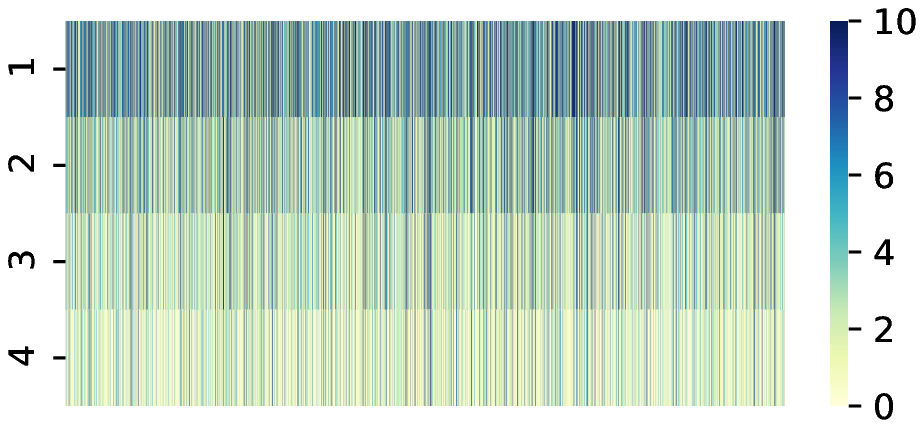}
    }
    \subcaptionbox{TDR (MRC-Rand)}{
        \includegraphics[width=0.23\textwidth]{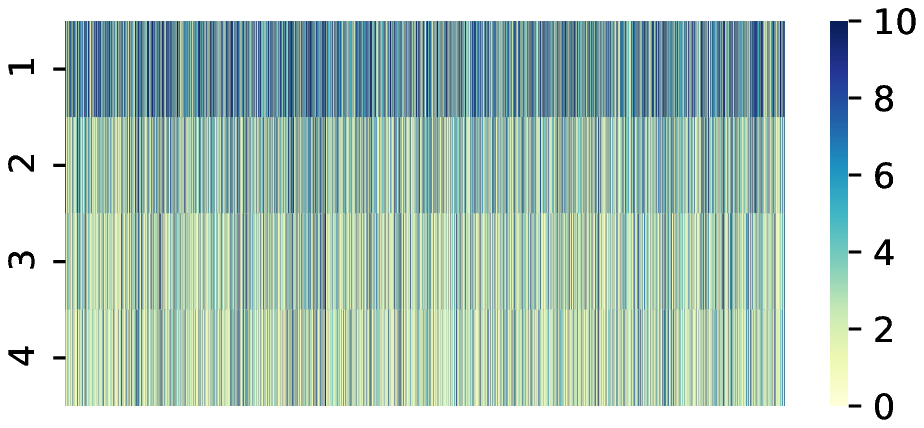}
    }

    \subcaptionbox{CDIE (MD3)}{
        \includegraphics[width=0.23\textwidth]{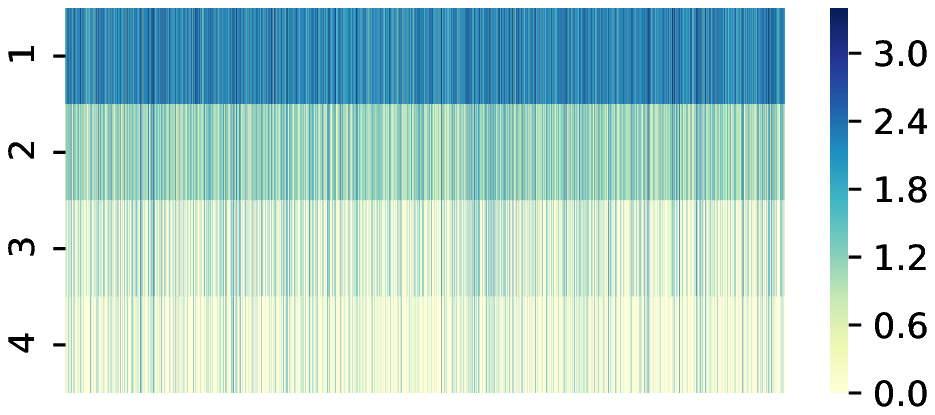}
    }
    \subcaptionbox{CDIE (MD3-Rand)}{
        \includegraphics[width=0.23\textwidth]{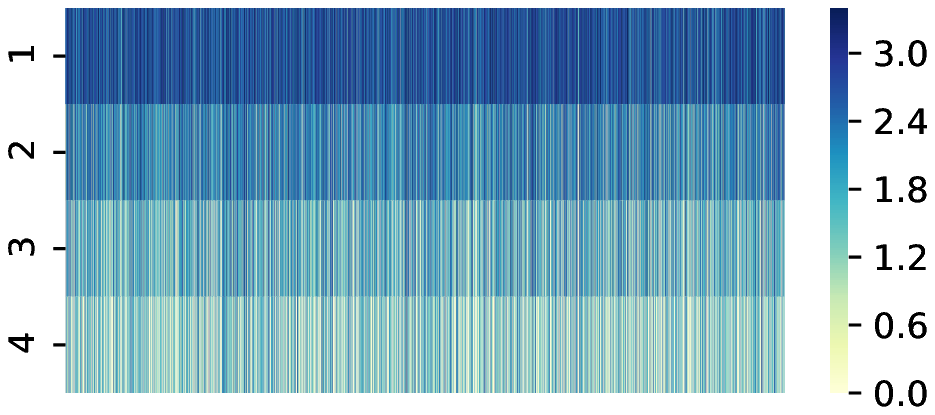}
    }
    \subcaptionbox{CDIE (MRC-Rand)}{
        \includegraphics[width=0.23\textwidth]{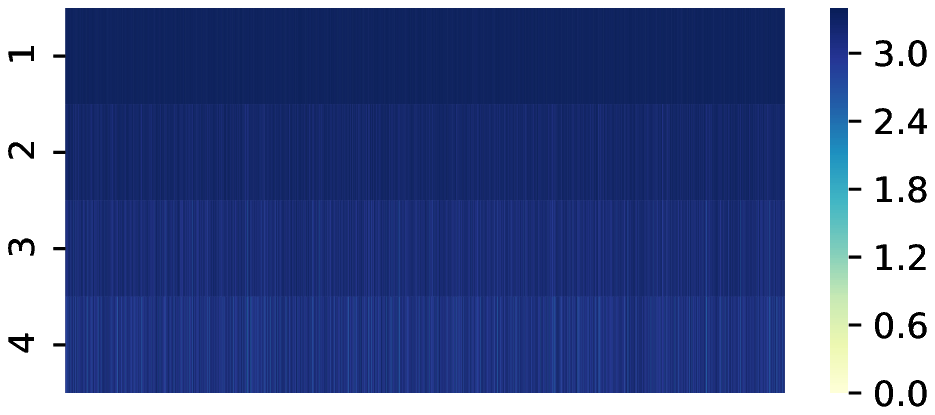}
    }

    \caption{Visualization of target document rank (TDR) and candidate documents information entropy (CDIE) dynamic changes. The ordinate represents the end of each dialogue turns. MD3-Rand and MRC-Rand denote DaLU or MRC with Rand policy.}
    \label{fig:doc-dynamics}
\end{figure*}

\begin{figure*}[h]
    \centering
    \includegraphics[width=0.9\textwidth]{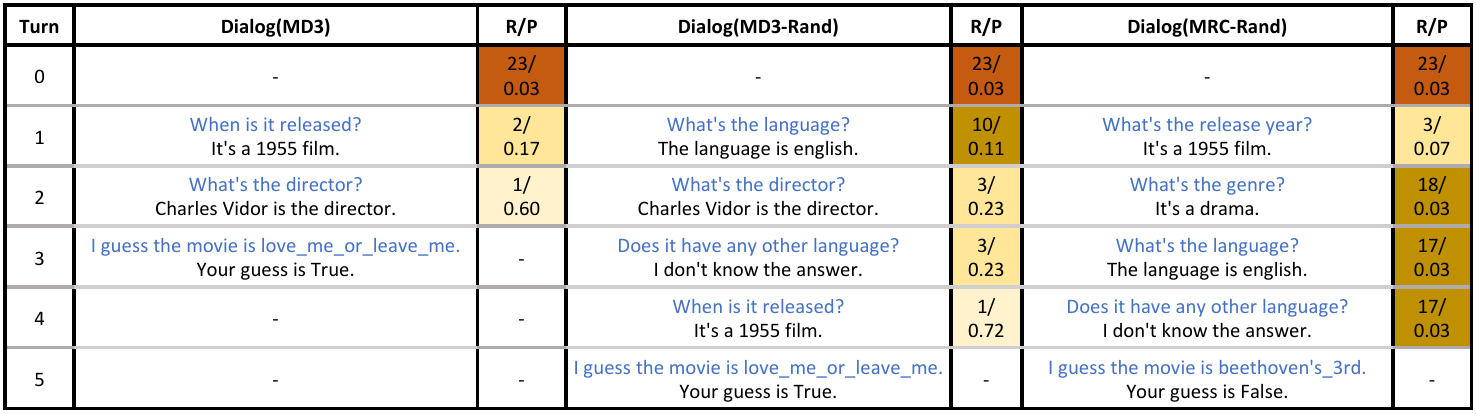}
    \caption{Examples of different dialogue models with the same candidate documents. R and P means target document rank and probability after each turn. The darker the color, the higher the rank, the lower the selection probability.}
    \label{fig:sample}
\end{figure*}

\subsection{Ablation Study}
\paragraph{Document Representation}
We use an Attribute-aware HAN (AaHAN) encoder for document representation, which accurately capture the key information for different attributes. By removing the attribute-aware mechanism, we only use the HAN encoder, which means each attribute share the same parameters. As shown in Table \ref{tab:AaHAN}, AaHAN has a significant improvement which demonstrates the importance of sufficient document knowledge representation for such dialogue.
\begin{table}[htb]
    \centering
    \small
    \begin{tabular}{l|lllll}
        \hline
        \textbf{Encoder} & \textbf{S1} & \textbf{S3} & \textbf{M} & \textbf{T} & \textbf{R} \\ \hline
        \textbf{AaHAN} & \textbf{0.83} &\textbf{ 0.97} & \textbf{0.90} & \textbf{3.42} & \textbf{1.73} \\
        HAN & 0.33 & 0.63 & 0.52 & 4.89 & 0.67 \\
        \hline
    \end{tabular}
    \caption{Dialogue test results for different encoder on document representation with candidate size of 32.}
    \label{tab:AaHAN}
\end{table}

\paragraph{Dialog Policy}
To demonstrate the necessary of attribute uncertainty $\gamma^t$ and attribute belief $\pi^t$ for dialog policy, we introduce two ablation tests on PM.
\begin{itemize}
    \item \texttt{w/o AU}: DaPO without attribute uncertainty $\gamma^t$.
    \item \texttt{w/o AB}: DaPO without attribute belief $\pi^t$.
\end{itemize}

As shown in Table \ref{tab:results}, both performance is significantly degraded over several metrics and different size of candidates. To make an accurately guess in shortest turns, we should not only consider the higher attribute uncertainty, but also the attribute belief.

\paragraph{Guess Threshold}
By adjusting the threshold $K$ which decides whether to make a guess, different policies can be obtained, as shown in the Table \ref{tab:doc-thre}. When the threshold $ K $ is large, it tends to make accurate guess and is not limited to the shortest dialogue turns. Otherwise, the top-1 document maybe not accurate, but the dialogue turns is shorter. This is a contradictory problem. We finally select threshold value of $0.5$ to obtain a sub-optimal policy.
\begin{table}[htb]
    \centering
    \small
    \begin{tabular}{l|lllll}
        \hline
        \textbf{K} & \textbf{S1} & \textbf{S3} & \textbf{M} & \textbf{T} & \textbf{R} \\ \hline
        \textbf{0.5} & 0.8298 & 0.9708 & 0.901 & \textbf{3.42} & \textbf{1.73} \\
        0.7 & 0.8756 & 0.9726 & 0.926 & 3.88 & 1.71 \\
        0.9 & \textbf{0.8774} & \textbf{0.9728} & \textbf{0.927} & 4.44 & 1.66 \\
        \hline
    \end{tabular}
    \caption{Dialogue test results for different document guess thresholds with candidate size of 32.}
    \label{tab:doc-thre}
\end{table}

\subsection{Analysis}
At the end of each dialogue, a sorting results can be obtained according to the Dialog-Level Doc Belief $p^t$. For all 5k simulated dialogs, we visualize the dynamic change of the target document rank (TDR), as shown in \ref{fig:doc-dynamics}-a. We can see that from top to bottom, as dialogue goes on, the color of the block is gradually lighter, which means TDR is gradually higher. In addition, the candidate documents information entropy (CDIE) calculated by the Dialog-Level Doc Belief $p^t$ can also be visualized, indicating uncertainty change of the selected document, as shown in \ref{fig:doc-dynamics}-b. From top to bottom, as dialogue goes on, the CDIE gradually decreases, which means the uncertainty of the guessed document become smaller. It illustrates that our model is interpretable.

In the comparison of several combinations, we can find that MD3 has significant advantages regardless of TDR and CDIE.

\subsection{Case Study}
Figure \ref{fig:sample} shows three dialogue samples between different models with the same candidates, and dynamic changes of the target rank and probability.

At the beginning of the dialogue, each document has the same probability to be guessed as the target. In the MD3 sample, the ``\textit{release year}'' and ``\textit{directed by}'' attributes are asked, so that the rank of the target document quickly rises to the first place, and the probability value increases to 0.6. As for the MRC-Rand sample, it doesn't ask the director, which has the greatest difference for candidates. And the probability of selecting the target is always low. This is due to the inaccuracy of the MRC model for attributes extraction and implied attribute information. It can be seen that MD3 is better than the other methods and more robust.

\section{Conclusions}
In this paper, we introduced a new multi-document driven dialogue task, and proposed a public benchmark dataset---GuessMovie. Besides, we investigated a multi-document driven dialogue model which can converse with the user and achieve the dialogue goal conditioned on both document engagement and user feedback. Additionally, although our model has significant advantages over several strong baselines. We hypothesize that there are several predefined attributes for documents and agent can only ask such questions. How to extend the dialogue to more scenarios with less restriction is a further research.

\section{Acknowledgments}
We thank the anonymous reviewers for their insightful comments. The research is supported in part by the Natural Science Foundation of China (Grant No. 62076032) and National Key Research and Development Program of China (Grant No. 2020YFF0305302).

\bibliography{ref}

\end{document}